\documentclass[manuscript,screen]{acmart}

\usepackage{xcolor}
\usepackage{amsmath}
\usepackage{booktabs}
\usepackage{multirow}
\usepackage{comment}
\usepackage{amsfonts}
\usepackage{float}

\usepackage{amsthm}
\theoremstyle{definition}
\newtheorem{definition}{Definition}
\newtheorem{example}{Example}

\usepackage[htt]{hyphenat}

\AtBeginDocument{%
  \providecommand\BibTeX{{%
    \normalfont B\kern-0.5em{\scshape i\kern-0.25em b}\kern-0.8em\TeX}}}

\setcopyright{acmcopyright}
\copyrightyear{2018}
\acmYear{2018}
\acmDOI{10.1145/1122445.1122456}

\acmConference[Woodstock '18]{Woodstock '18: ACM Symposium on Neural
  Gaze Detection}{June 03--05, 2018}{Woodstock, NY}
\acmBooktitle{Woodstock '18: ACM Symposium on Neural Gaze Detection,
  June 03--05, 2018, Woodstock, NY}
\acmPrice{15.00}
\acmISBN{978-1-4503-XXXX-X/18/06}

\begin{document}

%%
%% The "title" command has an optional parameter,
%% allowing the author to define a "short title" to be used in page headers.
\title{Assessment Modeling: Fundamental Pre-training Tasks for Interactive Educational Systems}

%%
%% The "author" command and its associated commands are used to define
%% the authors and their affiliations.
%% Of note is the shared affiliation of the first two authors, and the
%% "authornote" and "authornotemark" commands
%% used to denote shared contribution to the research.
\author{Youngduck Choi}
\affiliation{%
  \institution{Riiid! AI Research}
  \city{Seoul}
  \country{Republic of Korea}}
\email{youngduck.choi@riiid.co}

\author{Youngnam Lee}
\affiliation{%
  \institution{Riiid! AI Research}
  \city{Seoul}
  \country{Republic of Korea}}
\email{yn.lee@riiid.co}

\author{Junghyun Cho}
\affiliation{%
  \institution{Riiid! AI Research}
  \city{Seoul}
  \country{Republic of Korea}}
\email{jh.cho@riiid.co}

\author{Jineon Baek}
\affiliation{%
  \institution{Riiid! AI Research}
  \city{Seoul}
  \country{Republic of Korea}}
\email{jineon.baek@riiid.co}

\author{Dongmin Shin}
\affiliation{%
  \institution{Riiid! AI Research}
  \city{Seoul}
  \country{Republic of Korea}}
\email{dm.shin@riiid.co}

\author{Hangyeol Yu}
\affiliation{%
  \institution{Riiid! AI Research}
  \city{Seoul}
  \country{Republic of Korea}}
\email{hangyeol.yu@riiid.co}

\author{Yugeun Shim}
\affiliation{%
  \institution{Riiid! AI Research}
  \city{Seoul}
  \country{Republic of Korea}}
\email{yugeun.shim@riiid.co}

\author{Seewoo Lee}
\affiliation{%
  \institution{Riiid! AI Research}
  \city{Seoul}
  \country{Republic of Korea}}
\email{seewoo.lee@riiid.co}

\author{Jonghun Shin}
\affiliation{%
  \institution{Riiid! AI Research}
  \city{Seoul}
  \country{Republic of Korea}}
\email{jonghun.shin@riiid.co}

\author{Chan Bae}
\affiliation{%
  \institution{Riiid! AI Research}
  \city{Seoul}
  \country{Republic of Korea}}
\email{chan.bae@riiid.co}

\author{Byungsoo Kim}
\affiliation{%
  \institution{Riiid! AI Research}
  \city{Seoul}
  \country{Republic of Korea}}
\email{byungsoo.kim@riiid.co}

\author{Jaewe Heo}
\affiliation{%
  \institution{Riiid! AI Research}
  \city{Seoul}
  \country{Republic of Korea}}
\email{jwheo@riiid.co}

%%
%% By default, the full list of authors will be used in the page
%% headers. Often, this list is too long, and will overlap
%% other information printed in the page headers. This command allows
%% the author to define a more concise list
%% of authors' names for this purpose.
\renewcommand{\shortauthors}{Youngduck, et al.}

%%
%% The abstract is a short summary of the work to be presented in the
%% article.

\begin{abstract}
Like many other domains in Artificial Intelligence (AI), there are specific tasks in the field of AI in Education (AIEd) for which labels are scarce and expensive, such as predicting exam score or review correctness.
A common way of circumventing label-scarce problems is pre-training a model to learn representations of the contents of learning items.
However, such methods fail to utilize the full range of student interaction data available and do not model student learning behavior.
To this end, we propose Assessment Modeling, a class of fundamental pre-training tasks for general interactive educational systems.
An assessment is a feature of student-system interactions which can serve as a pedagogical evaluation. 
Examples include the correctness and timeliness of a student's answer.
Assessment Modeling is the prediction of assessments conditioned on the surrounding context of interactions.
Although it is natural to pre-train on interactive features available in large amounts, limiting the prediction targets to assessments focuses the tasks' relevance to the label-scarce educational problems and reduces less-relevant noise.
While the effectiveness of different combinations of assessments is open for exploration, we suggest Assessment Modeling as a first-order guiding principle for selecting proper pre-training tasks for label-scarce educational problems.
\end{abstract}


%%
%% Keywords. The author(s) should pick words that accurately describe
%% the work being presented. Separate the keywords with commas.

%%
%% This command processes the author and affiliation and title
%% information and builds the first part of the formatted document.
\maketitle

\section{Introduction}

\begin{figure*}
    \centering
    \includegraphics[width=0.8\textwidth]{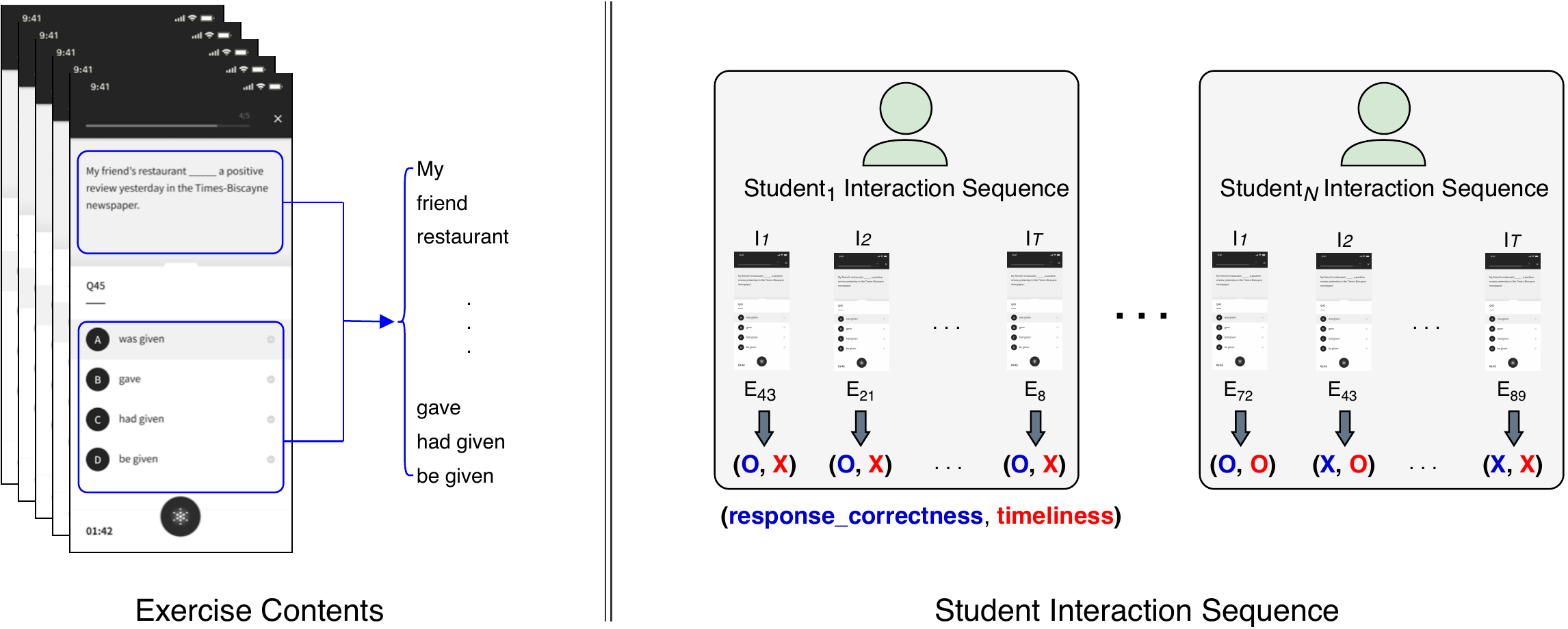}
    \caption{Comparison between content-based and interaction-based approaches.
    Content-based approaches learn representations of contents in learning items (e.g. exercises).
    On the other hand, interaction-based approaches model student learning behaviors in interactive educational systems.}
    \label{fig:content-interaction}
\end{figure*}

Interactive Educational Systems (IESs) have been developed rapidly in recent years to address the issue of quality and affordability in education.
IESs automatically collect observations of student behaviors at scale, and thus can power data-driven approaches for many Artificial Intelligence in Education (AIEd) tasks.
However, there are important tasks where a label-scarce problem prevents relevant models from attaining their full potential.
For instance, information about exam scores and grades are essential to understanding a student's educational progress and is a key factor affecting social outcomes.
However, unlike interactive features automatically collected by IESs, obtaining the labels is costly as they are often generated outside the IESs.
Other examples of scarce labels include data on course dropout and review correctness.
While this data is automatically recorded by IESs, they tend to be few in number as the events occur sporadically in practice.

Pre-train/fine-tune paradigm is a common way of circumventing the label-scarce problem that has been actively explored in the machine learning community.
In this paradigm, a model is first pre-trained in an unsupervised auxiliary task for which data is abundant.
Then, the model is slightly modified to match the main task and fine-tuned with possibly scarce data.
This approach has seen success in other subfields of AI including Natural Language Processing (NLP), computer vision, and motion planning \cite{devlin2018bert,studer2019comprehensive,schneider2019wav2vec}.
Following this line of inquiry, content-based pre-train/fine-tune methods \cite{huang2017question, sung2019improving, yin2019quesnet} have been studied in AIEd community.
However, student interactions are not considered and the work was limited to capturing the content of learning materials.
Accordingly, they do not make use of the information carried by the learning behavior of students using IESs.

In this paper, we propose Assessment Modeling, a class of fundamental pre-training tasks for general IESs.
Here, an assessment is any feature of student-system interactions which can act as a criterion for pedagogical evaluation.
Examples of assessments include the correctness and timeliness of a student response to a given exercise.
While there is a wide range of interactive features available, we narrow down the prediction targets to assessments to focus on the information most relevant to label-scarce educational problems.
Inspired by the recent success of bidirectional representations in NLP domain \cite{devlin2018bert}, we develop an assessment model using a deep bidirectional Transformer \cite{vaswani_2017} encoder.
In the pre-training phase, we randomly select a portion of entries in a sequence of interactions and mask the corresponding assessments.
Then, we train a deep bidirectional Transformer encoder-based assessment model to predict the masked assessments conditioned on the surrounding interactions.
After pre-training, we replace the last layer of the model with a layer corresponding to each label-scarce educational task, and all parameters of the model are then fine-tuned to the task.
To the best of our knowledge, this is the first work investigating appropriate pre-training methods for predicting educational features from student-system interactions.

We empirically evaluate the use of Assessment Modeling as pre-training tasks.
Our experiments are conducted on \emph{EdNet} \cite{choi2020ednet}, a large-scale dataset collected by an active mobile education application, \emph{Santa}, which has more than 131M response data points from around 780K students gathered since 2016.
The results show that Assessment Modeling provides a substantial performance improvement in label-scarce AIEd tasks.
In particular, we obtain improvements of 13.34\% mean absolute error and 4.26\% area under the receiving operating characteristic curve from the previous state-of-the-art model for exam score and review correctness prediction, respectively.

\section{Related Works}
Pre-training is the act of training a model to perform an unsupervised auxiliary task before using the trained model to perform the supervised main task \cite{erhan2010does}.
Pre-training has been shown to enhance the performance of models in various fields including NLP \cite{devlin2018bert,liu2019roberta,lan2019albert,yang2019xlnet,clark2020electra,brown2020language}, Computer Vision \cite{studer2019comprehensive,chen2020generative} and Speech Recognition \cite{schneider2019wav2vec}.
Pre-training techniques have been also applied to educational tasks with substantial performance improvements. 
For example, \cite{hunt2017transfer} predicts whether a student will graduate or not based on students’ general academic information such as SAT/ACT scores or courses taken during college.
They predict the graduation of 465 engineering students by first pre-training on the data of 6834 students in other departments using the TrAdaBoost algorithm \cite{dai2007boosting}.
\cite{ding2019transfer} suggests two transfer learning methods, Passive-AE transfer and Active-AE transfer, to predict student dropout in Massive Open Online Courses (MOOCs).
Their experimental results show that both methods improve the prediction accuracy, with Passive-AE transfer more effective for transfer learning across the same subject and Active-AE transfer more effective for transfer learning across different subjects. 

Most of the pre-training methods used in interactive educational system are NLP tasks with training data produced from learning materials.
For example, the short answer grading model suggested in \cite{sung2019improving} uses a pre-trained BERT model to ameliorate the limited amount of student-answer pair data.
They took a pre-trained, uncased BERT-base model and fine-tuned it on the ScientsBank dataset and two psychology domain datasets.
The resulting model outperformed existing grading models.
Test-aware Attention-based Convolutional Neural Network (TACNN) \cite{huang2017question} is a model that utilizes the semantic representations of text materials (document, question and options) to predict exam question difficulty (i.e. the percentage of examinees with wrong answer for a particular question). 
TACNN uses pre-trained word2vec embeddings \cite{mikolov2013distributedWORD2VEC} to represent word tokens. 
By applying convolutional neural networks to the sequence of text tokens and an attention mechanism to the series of sentences, the model quantifies the difficulty of the question. 
QuesNet \cite{yin2019quesnet} is a question embedding model pre-trained with the context information of question data. 
Since existing pre-training methods in NLP are unsuited for heterogeneous data such as images and metadata in questions, the authors suggest the Holed Language Model (HLM), a pre-training task, in parallel to BERT’s masking language model. 
HLM differs from BERT's task, however, because it predicts each input based on the values of other inputs aggregated in the Bi-LSTM layer of QuesNet, while BERT masks existing sequences at random. 
Also, QuesNet introduces another task called Domain-Oriented Objective (DOO), which is the prediction of the correctness of the answer supplied with the question, to capture high-level logical information. 
QuesNet adds a loss for each of HLM and DOO to serve as its final training loss.
Compared to other baseline models, QuesNet shows the best performance in three downstream tasks: knowledge mapping, difficulty estimation, and score prediction.

\section{Assessment Modeling}

\subsection{Formal Definition of Assessment Modeling}

Recall that Knowledge Tracing is the task of modeling a student’s knowledge state based on the history of their learning activities.
Although Knowledge Tracing is widely considered a fundamental task in AIEd and has been studied extensively, there is no precise definition in the literature. 
In this subsection, we first define Knowledge Tracing in a form that is quantifiable and objective for a particular IES design.
Subsequently, we introduce a definition of Assessment Modeling that addresses the educational values of the label being predicted.

A learning session in IES consists of a series of interactions $[I_{1}, \dots, I_{T}]$ between a student and the system, where each interaction $I_{t} = \{f_{t}^{1}, \dots, f_{t}^{n}\}$ is represented as a set of features $f_{t}^{k}$ automatically collected by the system.
The features represent diverse aspects of learning activities provided by the system, including the exercises or lectures being used, and the corresponding student actions.
Using the same notation, we define Knowledge Tracing and Assessment Modeling as follows:

\begin{definition}
\label{def:kt}
\emph{Knowledge Tracing} is the task of predicting a feature $f_{t}^{k}$ of the student in the $t$'th interaction $I_t$ given the sequence of interactions $[I_1, \dots, I_T]$.
That is, the prediction of
\begin{equation}
\label{eq:kt}
p(f_{t}^{k} | \{ I_1, \dots, I_T \} \setminus \Gamma(f_{t}^{k}))
\end{equation}
for some $\Gamma$, where $\Gamma(f)$ is the set of features that should be masked when the feature $f$ is guessed.
This is to mask input features not available at prediction time, so that the model does not cheat while predicting $f$.
\end{definition}

This definition is compatible with prior uses of the term in works on Knowledge Tracing models \cite{piech2015deep,zhang_2017,liu_2019,pandey2019self}. 
Although a common set-up of Knowledge Tracing models is to predict a feature conditioned on only past interactions, we define Knowledge Tracing as a prediction task that can also be conditioned on future interactions to encompass the recent successes of bi-directional architectures in Knowledge Tracing \cite{lee2019creating}.

\begin{example}[\textbf{Knowledge Tracing}]
\label{ex:kt}
A typical instance of a Knowledge Tracing task might be response correctness prediction,  where the interaction $I_i = \{e_i, r_i\}$ consists of an exercise $e_i$ given to a student, and the correctness $r_i$ of the student's corresponding response \cite{piech2015deep,zhang_2017,liu_2019,pandey2019self,lee2019creating}.
In this setup, only the response correctness $r_T$ of the last interaction $I_T$ is predicted and the features related to $r_T$ are masked.
Following our definition of Knowledge Tracing, the task can be extended further to predict diverse interactive features such as:
\begin{itemize}
    \item \emph{offer\_selection}: Whether a student accepts studying the offered learning items.
    \item \emph{start\_time}: The time a student starts to solve an exercise.
    \item \emph{inactive\_time}: The duration for which a student is inactive in a learning session.
    \item \emph{platform}: Whether a student responds to each exercise on a web browser or a mobile app.
    \item \emph{payment}: Whether a student purchases paid services.
    \item \emph{event} : Whether a student participates in application events.
    \item \emph{longest\_answer}: Whether a student selected the answer choice with the longest description.
    \item \emph{correctness}: Whether a student responds correctly to a given exercise.
    \item \emph{timeliness}: Whether a student responds to each exercise under the time limit recommended by domain experts.
    \item \emph{course\_dropout} : Whether a student drops out of the entire class.
    \item \emph{elapsed\_time}: The duration of time a student takes to solve a given exercise.
    \item \emph{lecture\_complete}: Whether a student completes studying a video lecture offered to them.
    \item \emph{review\_correctness} : Whether a student responds correctly to a previously solved exercise.
\end{itemize}
\end{example}

In the aforementioned example, features like \emph{correctness} and \emph{timeliness} directly evaluate the educational values of a student interaction,
while it is somewhat debatable whether \emph{platform} and \emph{longest\_answer} are also capable of addressing such qualities. 
Accordingly, we define \emph{assessments} and \emph{Assessment Modeling} as the following:

\begin{definition}
\label{def:am}
An \emph{assessment} $a_{t}^{k}$ of the $t$'th interaction $I_t$ is a feature of $I_t$ which can act as a criterion for pedagogical evaluation. 
The collection $A_{t} = \{a_{t}^{1}, \dots, a_{t}^{m}\}$ of assessments is a subset of the available features $\{f_{t}^{1}, \dots, f_{t}^{n}\}$ of $I_t$.
\emph{Assessment Modeling} is the prediction of assessment $a_{t}^{k}$ for some $k$ 
from the interactions $[I_1, \dots, I_T]$.
That is, the prediction of
\begin{equation}
\label{eq:am}
p(a_{t}^{k} | \{ I_1, \dots, I_T \} \setminus \Gamma(a_{t}^{k})).
\end{equation}
\end{definition}

\begin{figure}[t]
\includegraphics[width=0.6\textwidth]{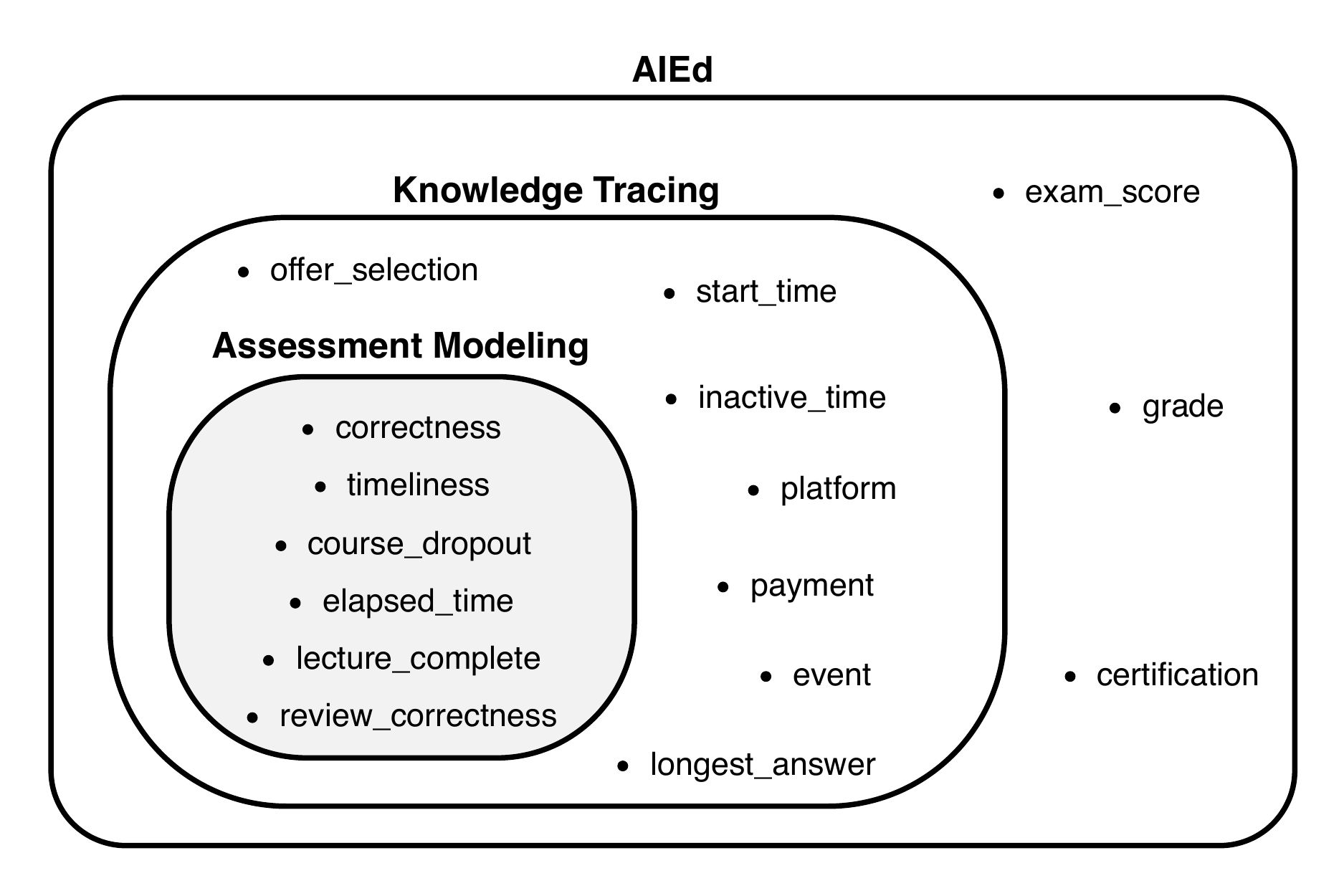}
\caption{The features predicted in general AIEd tasks, Knowledge Tracing and Assessment Modeling.
Assessment Modeling predicts the distribution of assessments, the subset of interactive features which can act as pedagogical evaluation criteria.
Note that predicting an exam score (\emph{exam\_score}), a grade (\emph{grade}) and whether a student will pass to get a certificate (\emph{certification}) are tasks outside Knowledge Tracing.}
\label{fig:am_def}
\end{figure}

\begin{example}[\textbf{Assessments}]
\label{ex:am}
Among the interactive features listed in Example \ref{ex:kt}, we consider \emph{correctness}, \emph{timeliness}, \emph{course\_dropout}, \emph{elapsed\_time}, \emph{lecture\_complete} and \emph{review\_correctness} to be assessments.
For example, \emph{correctness} is an assessment as whether a student responded to each exercise correctly provides strong evidence regarding the student's mastery of concepts required to solve the exercise.
Also, \emph{timeliness} serves as an assessment since a student solving given exercise within the recommended time limit is expected to be proficient in skills and knowledge necessary to answer the exercise.
Figure \ref{fig:am_def} depicts the relationship between assessments and general Knowledge Tracing features.
\end{example}

\subsection{Assessment Modeling as Pre-training Tasks}

In this subsection, we provide examples of important yet scarce educational features and argue that Assessment Modeling enables effective prediction of such features.

\begin{example}[\textbf{Non-Interactive Educational Features}]
\label{ex:non-interactive}
In many applications, IES is often integrated as part of a larger learning process.
Accordingly, the ultimate evaluation of the learning process is mostly done independently from the IES.
For example, academic abilities of students are measured by course grades or standardized exams, and the ability to perform a complicated job or task is certified by professional certificates.
Such labels are considered essential due to the pedagogical and social needs for consistent evaluations of student ability.
However, obtaining these labels is often challenging due to their scarcity compared to that of features automatically collected from student-system interactions.
We give the following examples (Figure \ref{fig:am_def}):
\begin{itemize}
    \item \emph{exam\_score}: A student's score on a standardized exam.
    \item \emph{grade}: A student's final grade in a course.
    \item \emph{certification}: Professional certifications obtained by completion of educational programs or examinations.
\end{itemize}
\end{example}

\begin{figure}[t]
\includegraphics[width=0.5\textwidth]{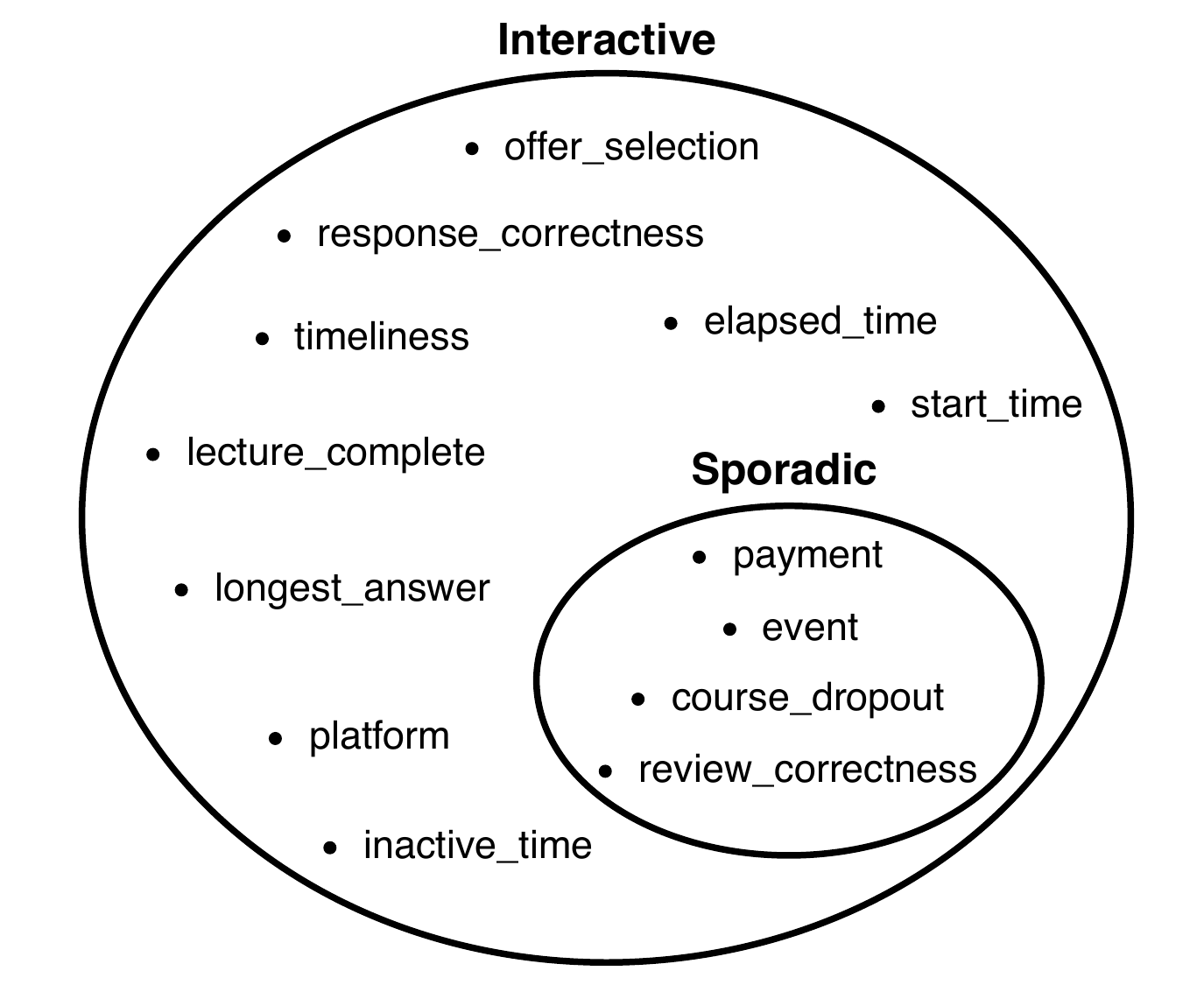}
\caption{Not all interactive features are collected equally often.
For example, \emph{payment}, \emph{event}, \emph{course\_dropout} and \emph{review\_correctness} are interactive features obtained more sporadically than other features.
Among the sporadic interactive features, we consider \emph{course\_dropout} and \emph{review\_correctness} as assessments.}
\label{fig:am_freq}
\end{figure}

\begin{example}[\textbf{Sporadic Assessments}]
\label{ex:sporadic}
All assessments are automatically collected by IESs, but some assessments are few in number as the corresponding events occur rarely in practice.
For example, it is natural for students to invest more time in learning new concepts than reviewing previously studied materials.
\emph{course\_dropout} and \emph{review\_correctness} are examples of \emph{sporadic assessments} (Figure \ref{fig:am_freq}).
\end{example}

To overcome the aforementioned lack of labels, we consider the pre-train/fine-tune paradigm that leverages data available in large amounts to aid performance in tasks where labels are scarce.
In this paradigm, a model is first trained in an auxiliary task relevant to the tasks of interest with label-scarce data.
Using the pre-trained parameters to initialize the model, the model is slightly modified to suit the task of interest, and then fine-tuned on the main tasks.
This approach has been successful in AI fields like NLP, computer vision and speech recognition \cite{devlin2018bert,studer2019comprehensive,schneider2019wav2vec}.
Following this template, existing methods in AIEd pre-train on the contents of learning materials, but such methods do not capture student behavior and only utilize a small subset of features available from the data.

\begin{figure}[t]
\includegraphics[width=0.6\textwidth]{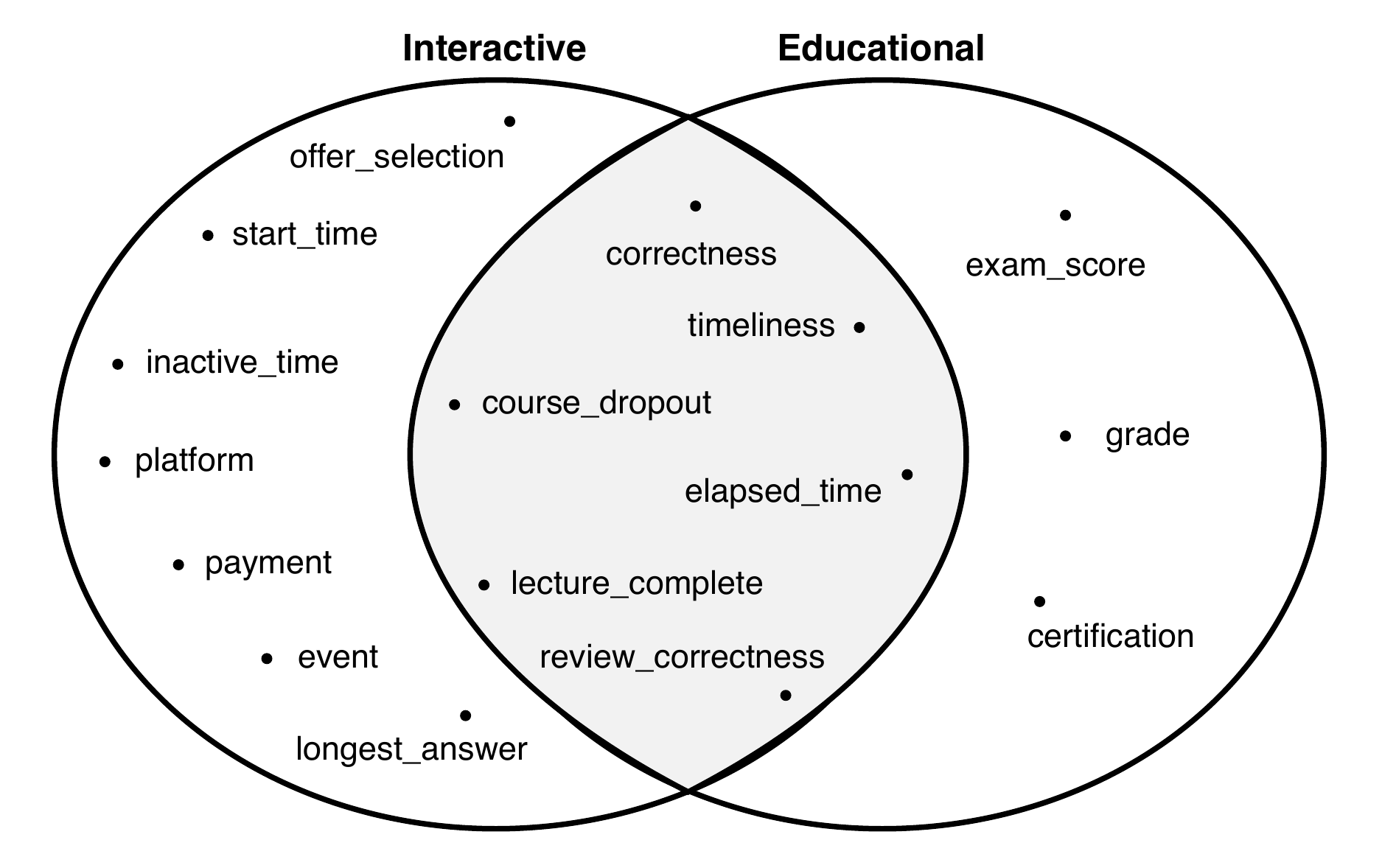}
\caption{Assessment Modeling is an effective pre-training method for label-scarce educational problems.
First, most assessments are available in large amounts as they are automatically collected from student-system interactions (Interactive).
Also, assessments are selected to be relevant to educational progress, narrowing down the scope of prediction targets and reducing noise irrelevant to the problems (Educational).}
\label{fig:am_alt}
\end{figure}

\begin{figure}[t]
\includegraphics[width=0.6\textwidth]{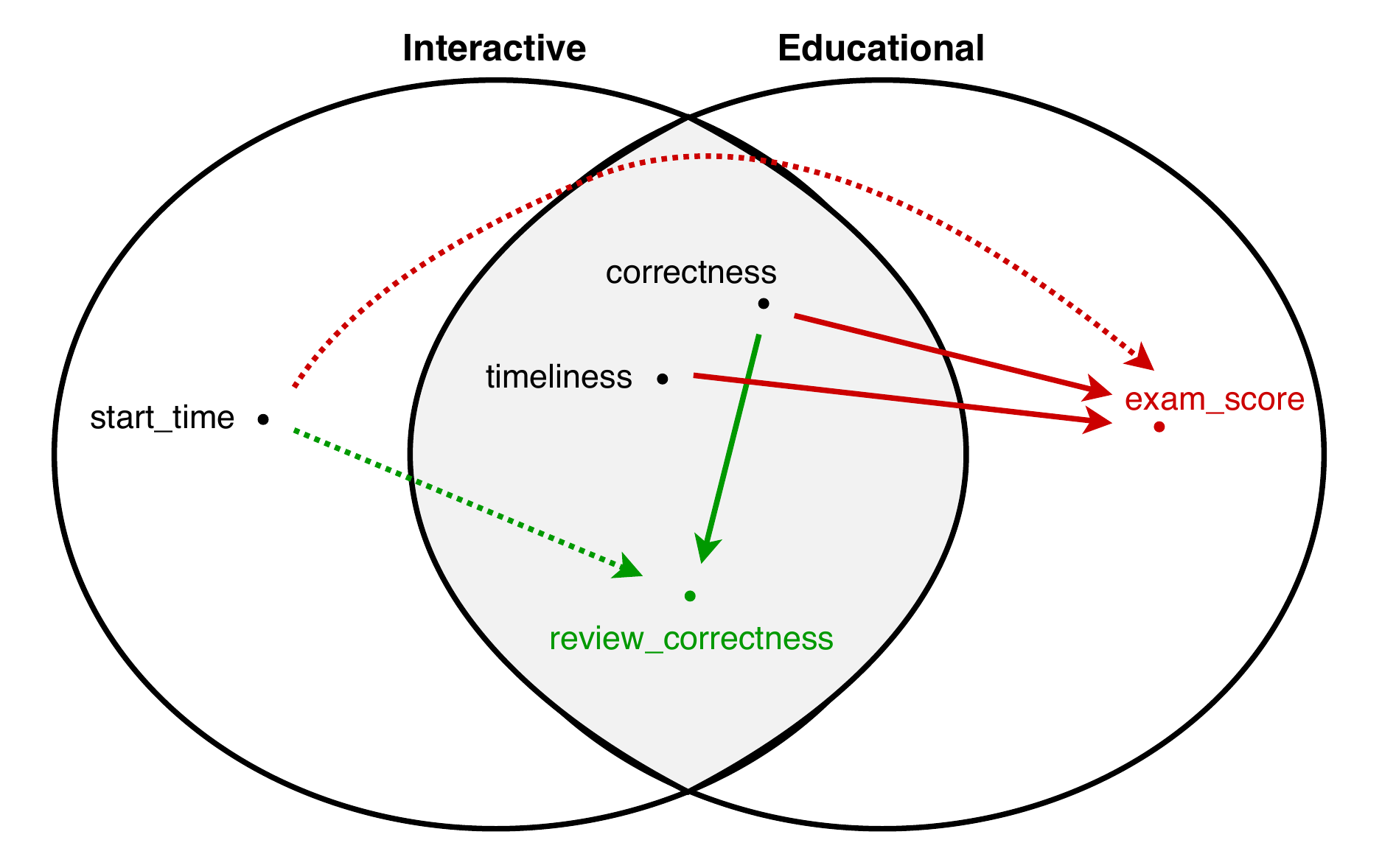}
\caption{Possible pre-train/fine-tune scenarios.
We may pre-train a model to predict \emph{start\_time}, \emph{correctness} and \emph{timeliness}, and then train it to estimate \emph{exam\_score} (red).
Likewise, a model pre-trained to predict \emph{start\_time} and \emph{correctness} can be trained to predict \emph{review\_correctness} (green).
However, pre-training to predict non-educational interactive features (non-assessment) like \emph{start\_time} is not effective in label-scarce educational problems (dotted line).}
\label{fig:am_pre_fine}
\end{figure}

Instead, one may pre-train on different features automatically collected by IESs (Figure \ref{fig:am_alt}).
However, training on every available feature is computationally intractable and may introduce irrelevant noise.
To this end, Assessment Modeling narrows down the prediction targets to assessments, the interactive features that also hold information on educational progress.
Since multiple assessments are available, a wide variety of pre-train/fine-tune pairs can be explored for effective Assessment Modeling (Figure \ref{fig:am_pre_fine}).
This raises the open-ended questions of which assessments to pre-train on for label-scarce educational problems and how to pre-train on multiple assessments.

\subsection{Assessment Modeling with Deep Bidirectional Transformer Encoder}

\begin{figure*}
    \centering
    \includegraphics[width=0.9\textwidth]{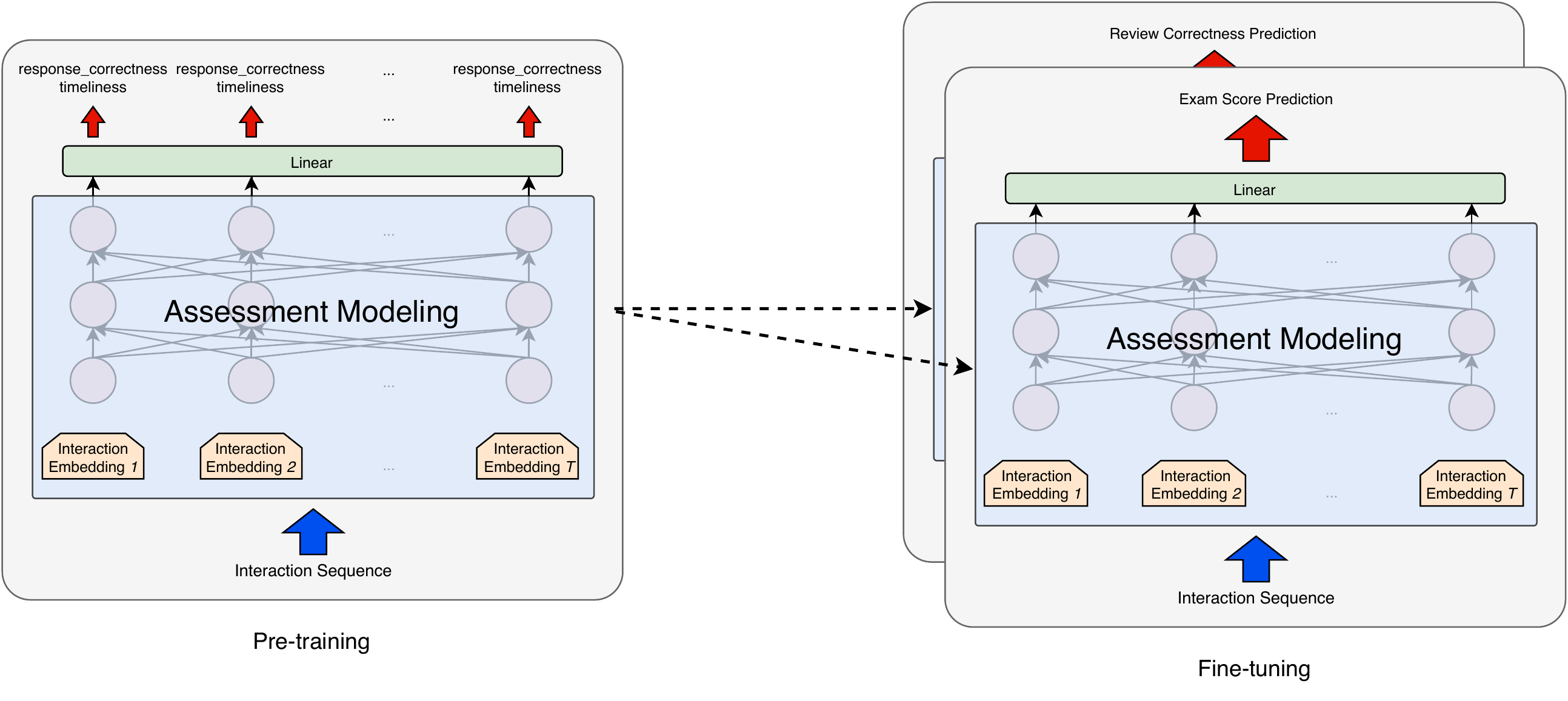}
    \caption{Proposed pre-train/fine-tune approach.
    In the pre-training phase, we train an assessment model to predict assessments conditioned on past and future interactions.
    After the pre-training phase, we fine-tune parameters in the model to predict labels in label-scarce educational tasks.}
    \label{fig:am_model}
\end{figure*}

While there are several possible options for the architecture of the assessment model, we adopt the deep bidirectional Transformer encoder proposed in \cite{devlin2018bert} for the following reasons.
First, \cite{pandey2019self} showed that the self-attention mechanism in Transformer \cite{vaswani_2017} is effective for Knowledge Tracing.
The Transformer-based Knowledge Tracing model proposed in \cite{pandey2019self} achieved state-of-the-art performance on several datasets.
Second, the deep bidirectional Transformer encoder model and pre-train/fine-tune method proposed in \cite{devlin2018bert} achieved state-of-the-art results on several NLP tasks.
While \cite{devlin2018bert} conducted experimental studies in the NLP domain, the method is also applicable to other domains with slight modifications.
Figure \ref{fig:am_model} depicts our proposed pre-train/fine-tune approach.
In the pre-training phase, we train a deep bidirectional Transformer encoder-based assessment model to predict assessments conditioned on past and future interactions.
After the pre-training phase, we replace the last layer of the assessment model with a layer appropriate for each label-scarce educational task and fine-tune parameters in the whole model to predict labels in the task.
We provide detailed descriptions of our proposed assessment model in the following subsections.

\subsubsection{Input Representation}
The first layer in the assessment model maps each interaction to an embedding vector.
First, we embed the following attributes:
\begin{itemize}
    \item \emph{exercise\_id}: We assign a latent vector unique to each exercise id.
    \item \emph{exercise\_category}: Each exercise has its own category tag that represents the type of the exercise.
    We assign a latent vector to each tag.
    \item \emph{position}: The relative position $t$ of the interaction $I_t$ in the input sequence.
    We use the sinusoidal positional encoding that is used in \cite{vaswani_2017}.
\end{itemize}  
As shown in Example \ref{ex:kt}, IES collects diverse interactive features that can potentially be used for Assessment Modeling.
However, not only using all possible interactive features for Assessment Modeling is computationally intractable, there is no guarantee that the best results on label-scarce educational tasks will be achieved when all the features are used.
For experimental studies, we narrow down the scope of interactive features to the ones available from an exercise-response pair, the simplest and widely-considered interaction in Knowledge Tracing. 
In particular, we embed the following interactive features:
\begin{itemize}
    \item \emph{correctness}: The value is 1 if a student response is correct and 0 otherwise. 
    We assign a latent vector corresponding to each possible value 0 and 1.
    \item \emph{elapsed\_time}: The time taken for a student to respond is recorded in seconds.
    We cap any time exceeding 300 seconds to 300 seconds and normalize it by dividing by 300 to have a value between 0 and 1.
    The elapsed time embedding vector is calculated by multiplying the normalized time by a single latent embedding vector.
    \item \emph{inactive\_time}: The time interval between adjacent interactions is recorded in seconds.
    We set maximum inactive time as 86400 seconds (24 hours) and any time more than that is capped off to 86400 seconds.
    Also, the inactive time is normalized to have a value between 0 and 1 by dividing the value by 86400.
    Similar to the elapsed time embedding vector, we calculate the inactive time embedding vector by multiplying the time by a single latent embedding vector.
    %\item \emph{timeliness}: The value is 1 if a student responds within a specified time limit and 0 otherwise. 
    %We assign a latent vector corresponding to each possible value 0 and 1.
\end{itemize}
Let $e_t$ be the sum of the embedding vectors of \emph{exercise\_id}, \emph{exercise\_category} and \emph{position}.
Likewise, let $c_t$, $et_t$ and $it_t$ be the embedding vectors of \emph{correctness}, \emph{elapsed\_time} and \emph{inactive\_time}, respectively.
Then, the representation of interaction $I_t$ is $e_t + c_t + et_t + it_t$.

\subsubsection{Masking}
Inspired by the masked language model proposed in \cite{devlin2018bert}, we use the following method to mask the assessments in a student interaction sequence.
First, we mask a fraction $M$ of interactions chosen uniformly at random.
If the $t$-th interaction is chosen, we replace embedding vectors of interactive features that are being predicted as targets with $mask$, a trainable vector that represents masking.
For instance, if the \emph{correctness} and \emph{elapsed\_time} are prediction targets for Assessment Modeling, $c_t + et_t$ is replaced with $mask$ and the embedding vector for interaction $I_t$ becomes $e_t + it_t + mask$.

\subsubsection{Model Architecture}
After the interactions are embedded and masked accordingly, they enter a series of Transformer encoder blocks, each consisting of a multi-headed self-attention layer followed by position-wise feed-forward layer.
Every layer has input dimension $d_{\mathrm{model}}$.
The first encoder block takes the sequence of interactions $I_1, \dots, I_T$ embedded in latent space and returns a series of vectors of the same length and dimension.
For all $2 \leq i \leq N$, the $i$'th block takes the output of the $(i-1)$'th block as input and returns the series of vectors accordingly.
We describe the architecture of each block as the following.

The multi-headed self-attention layer takes a series of vectors, $X_1, \dots, X_T$.
Each vector is projected to latent space by projection matrices $W^{Q}, W^{K} \in \mathbb{R}^{d_{\mathrm{model}} \times d_{K}}$ and $W^{V} \in \mathbb{R}^{d_{\mathrm{model}} \times d_{V}}$: 
\begin{align}
\label{eqn:attn-proj}
\begin{split}
Q = [q_1, \dots, q_T]^{T} & = X W^Q \\
K = [k_1, \dots, k_T]^{T} & = X W^K \\
V = [v_1, \dots, v_T]^{T} & = X W^V
\end{split}
\end{align}
Here $X = [X_1, \dots, X_T]^{T}$, and each $q_i$, $k_i$ and $v_i$ are the query, key and value of $X_i$, respectively.
The output of the self-attention is then obtained as a weighted sum of values with coefficients determined by the dot products between queries and keys:
\begin{align}
\label{eqn:attn-softmax}
\mathrm{Attention}(X) = \mathrm{Softmax}\left(\frac{QK^{T}}{\sqrt{d_k}}\right) V
\end{align}
Models with self-attention layers often use multiple heads to jointly attend information from different representative subspaces. 
Following this, we apply attention $h$ times to the same query-key-value entries with different projection matrices for output:
\begin{align}
\label{eqn:multihead}
\mathrm{Multihead}(X) = \mathrm{concat}(\mathrm{head}_{1}, \dots, \mathrm{head}_{T}) W^O
\end{align}
Here, each $\mathrm{head}_{i}$ is equal to the output of self-attention in Equation \ref{eqn:attn-softmax} with corresponding projection matrices $W_i^Q$, $W_i^K$ and $W_i^V$ in Equation \ref{eqn:attn-proj}.
We use the linear map $W^O$ to aggregate each attention result.
After we compute the resulting value in Equation \ref{eqn:multihead}, we apply point-wise feed-forward layer to add non-linearity to the model.
Also, we apply the skip connection \cite{he2016deep} and layer normalization \cite{ba2016layer} to the output of the feed-forward layer.

Assume that the last encoder block returns the sequence $H = [H_1, \dots, H_T]^T$ of vectors.
For pre-training, the predictions to features in $i$'th timestep are made by applying a linear layer with the proper activation function to $H_i$.
We consider four interactive features as prediction targets: \emph{correctness}, \emph{timeliness}, \emph{elapsed\_time} and \emph{inactive\_time}.
If the prediction target is \emph{correctness} or \emph{timeliness}, the sigmoid activation function is applied to the linear layer output.
If \emph{elapsed\_time} or \emph{inactive\_time} is the prediction target, the final output of the model is the output of the linear layer.
The overall loss is defined to be
\begin{align}
\label{eqn:loss}
\mathcal{L} = \sum_{t=1}^{T} m_{t}\mathcal{L}_{t}
\end{align}
where $\mathcal{L}_{t}$ is the sum of binary-cross-entropy (resp. mean-squared-error) losses if the prediction target is \emph{correctness} or \emph{timeliness} (resp. \emph{elapsed\_time} or \emph{inactive\_time}).
The value $m_{t}$ is a flag that represents whether the $t$-th exercise is masked ($m_{t}=1$) or not ($m_{t}=0$).
The input embedding layer and encoder blocks are shared over pre-training and fine-tuning. 
For fine-tuning, we replace the linear layers applied to each $H_i$ in pre-training with a single linear layer that combines all the entries of $H$ to fit the output to label-scarce educational downstream tasks.

\section{Experiments} \label{sec:experiments}

\subsection{Label-Scarce Educational Tasks}
We apply Assessment Modeling to exam score (a non-interactive educational feature) and review correctness (a sporadic assessment) predictions.

\subsubsection{Exam Score Prediction}
Exam score prediction is the estimation of student's scores in standardized exams, such as the TOEIC and the SAT, based on the student's interaction history with IES.
Exam score prediction is one of the most important tasks of AIEd, as standardized assessment is crucial for both the students and IES. 
Because a substantial amount of human effort is required to develop or take the tests, the number of data points available for exam score prediction is considerably fewer than that of student interactions automatically collected by IES.
By developing a reliable exam score prediction model, a student's universally accepted score can be estimated by IES with considerably less effort. 
Exam score prediction differs from response correctness prediction because standardized tests are taken in a controlled environment with specific methods independent of IES. 

\subsubsection{Review Correctness Prediction}
Assume that a student incorrectly responds to an exercise $e_{\mathrm{rev}}$ and receives corresponding feedback.
The goal of review correctness prediction is to predict whether a student will be able to respond to the exercise $e_{\mathrm{rev}}$ correctly if they encounter the exercise again.
The significance of this AIEd task is that it can assess the educational effect of an exercise to a particular student in a specific situation.
In particular, the correctness probability estimated by this task represents the student's expected marginal gain in knowledge as they go through some learning process. 
For example, if the correctness probability is high, it is likely that the student will obtain relevant knowledge in the future even if their initial response was incorrect.

\subsection{Dataset}
We use the public \emph{EdNet} \cite{choi2020ednet} dataset obtained from \emph{Santa}, a mobile AI tutoring service for TOEIC Listening and Reading Test preparation.
The test consists of two timed sections named Listening Comprehension (LC) and Reading Comprehension (RC) with a total of 100 exercises, and 4 and 3 parts, respectively. 
The final test score ranges from 10 to 990 in steps of 5. 
Once a student solves each exercise, \emph{Santa} provides educational feedback to their responses including explanations and commentaries on exercises.
\emph{EdNet} is the collection of student interactions of multiple-choice exercises, which contains more than 131M response data points from around 780K students gathered over the last four years.
The main features of the student-exercise interaction data consists of six columns: student id, exercise id, exercise part, student response, received time and time taken.
The student (resp. exercise) ID identifies each unique student (resp. exercise).
The student response is a student's answer choice for the given exercise.
Exercise part is the part of the exam that the exercise belongs to.
Finally, the absolute time when the student received the exercise and the time taken by the student to respond are recorded.

\begin{figure}[t]       
    \fbox{\includegraphics[width=0.2\textwidth]{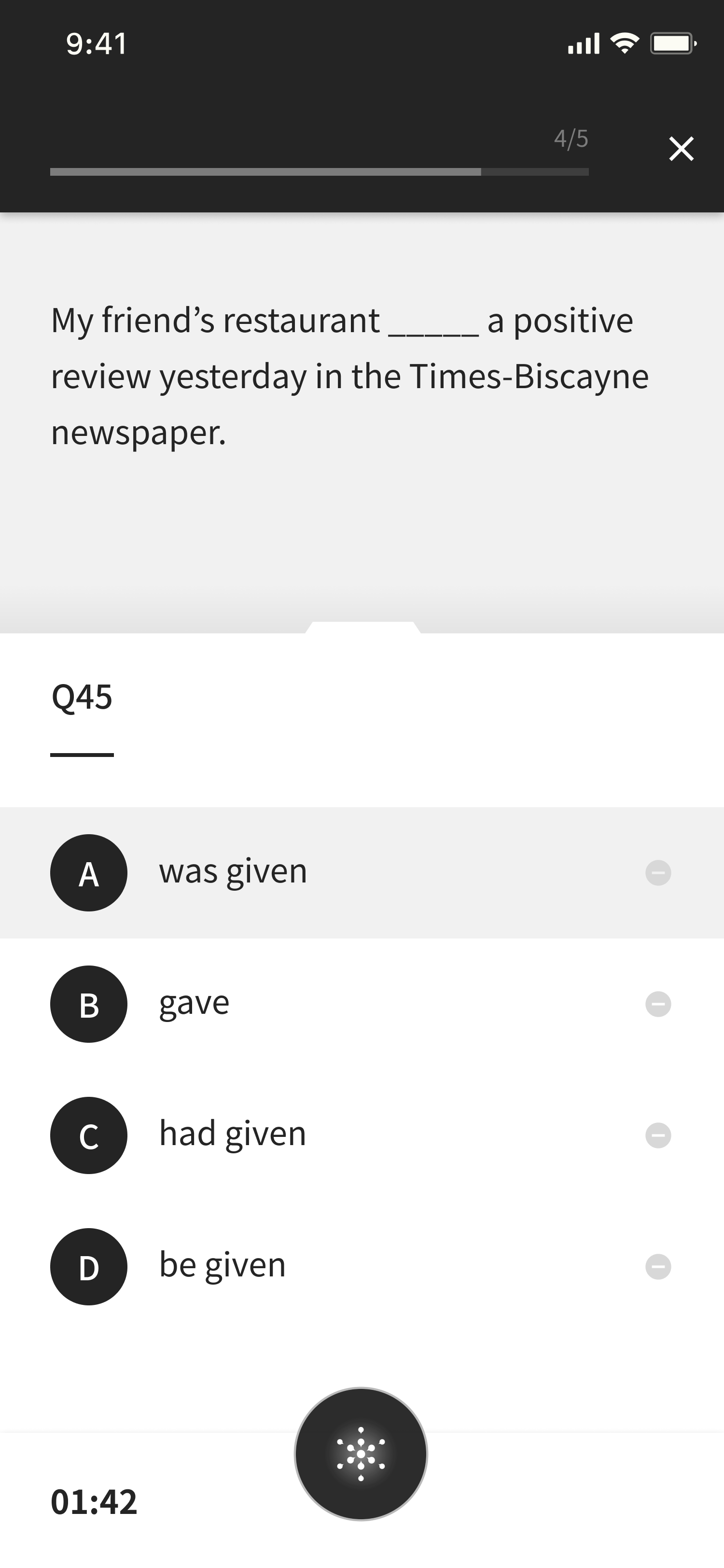}}   
    \hspace{30px}
    \fbox{\includegraphics[width=0.2\textwidth]{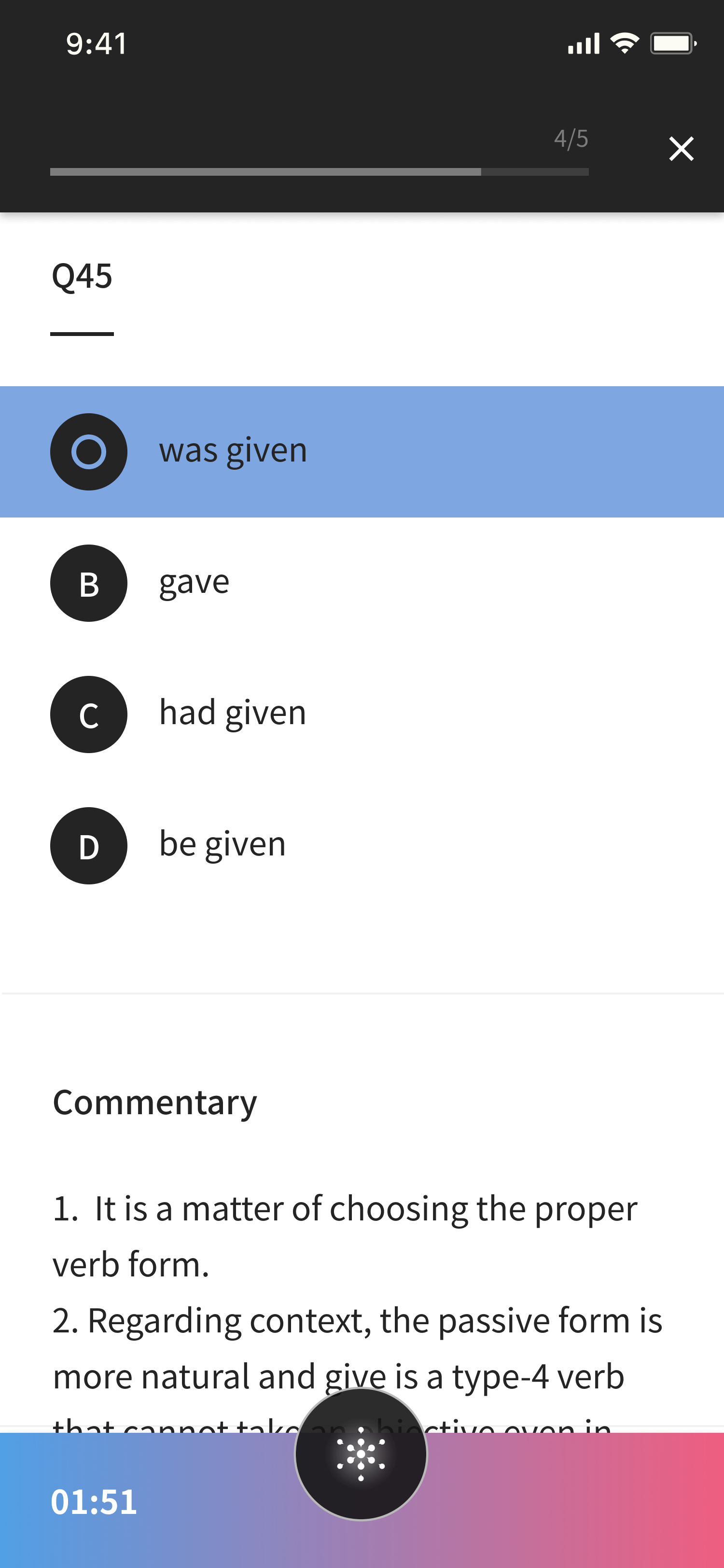}}
    \caption{Student Interface of \emph{Santa}}
    \label{fig:ui}
\end{figure}

\subsubsection{Dataset for Pre-training}
For pre-training, we first reconstruct the interaction timeline of each student by gathering the responses of a specific student in increasing chronological order.
For each interaction $I_t$, $\emph{correctness}_t$ is recorded as 1 if the student's answer is correct and 0 otherwise. $\emph{timeliness}_t$ is recorded as 1 if the student responded under the time limits recommended by TOEIC experts (Table \ref{tab:timelimit}) and 0 otherwise. $\emph{elapsed\_time}_t$ is the time taken for the student to respond recorded in seconds, and $\emph{inactive\_time}_t$ is the time interval between the current and previous interactions recorded in seconds.
We exclude the interactions of students involved in any of the label-scarce educational downstream tasks for pre-training to preemptively avoid data leakage. 
After processing, the data consists of 414,375 students with a total of 93,121,528 interactions.

\begin{table}[h]
\centering
\caption{Time Limits}
\begin{tabular}{c|cccc|ccc}
\toprule
Part            & \multicolumn{4}{c|}{ 1 $\sim$ 4} & 5  & 6  & 7  \\  
\hline
Time limit (sec)& \multicolumn{4}{c|}{audio duration + 8} & 25 & 50 & 55 \\
\bottomrule
\end{tabular}
\label{tab:timelimit}
\end{table}

\subsubsection{Dataset for Label-Scarce Educational Tasks}
For exam score prediction, we aggregate the real TOEIC scores reported by students of \emph{Santa}.
The reports are scarce in number because a student has to register, take the exam and report the score at their own expense. 
To collect this data, \emph{Santa} offered a small reward to students in exchange for reporting their score.
A total of 2,594 score reports were obtained over a period of 6 months, which is considerably fewer than the number of exercise responses. 
For our experiment, we divide the data into five splits, and use 3/5, 1/5 and 1/5 of the data as training, validation and test set, respectively.

For review correctness prediction, we look over each student's timeline and find exercises that have been solved at least twice.
That is, if an exercise $e$ appears more than once in a student interaction sequence $I_1 = (e_1, r_1), I_2 = (e_2, r_2), \dots, I_T = (e_T, r_T)$, we find the first two interactions $I_i$ and $I_j$ ($i < j$) with the same exercise $e$.
The sequence of interactions $I_{i+1}, \cdots, I_{j-1}$ and $e_j$ are taken as input, and $\emph{correctness}_j$ is taken as the label.
The total of 4,540 labeled sequences which are not appeared in the pre-training dataset are generated after pre-processing.
Similar to the case in exam score prediction, we divide the data into five splits, and use 3/5, 1/5 and 1/5 of the data as training, validation and test set, respectively.

\begin{figure}[t]       
    \fbox{\includegraphics[width=0.25\textwidth]{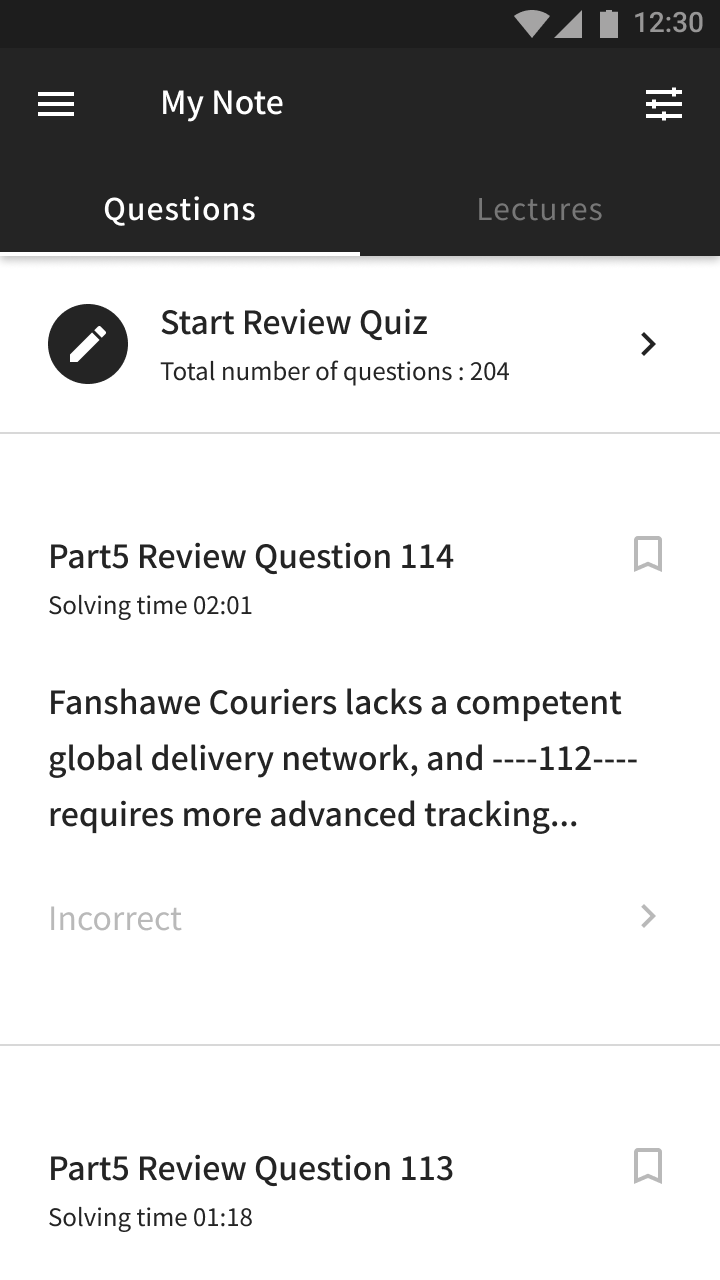}}   
    \hspace{30px}
    \fbox{\includegraphics[width=0.25\textwidth]{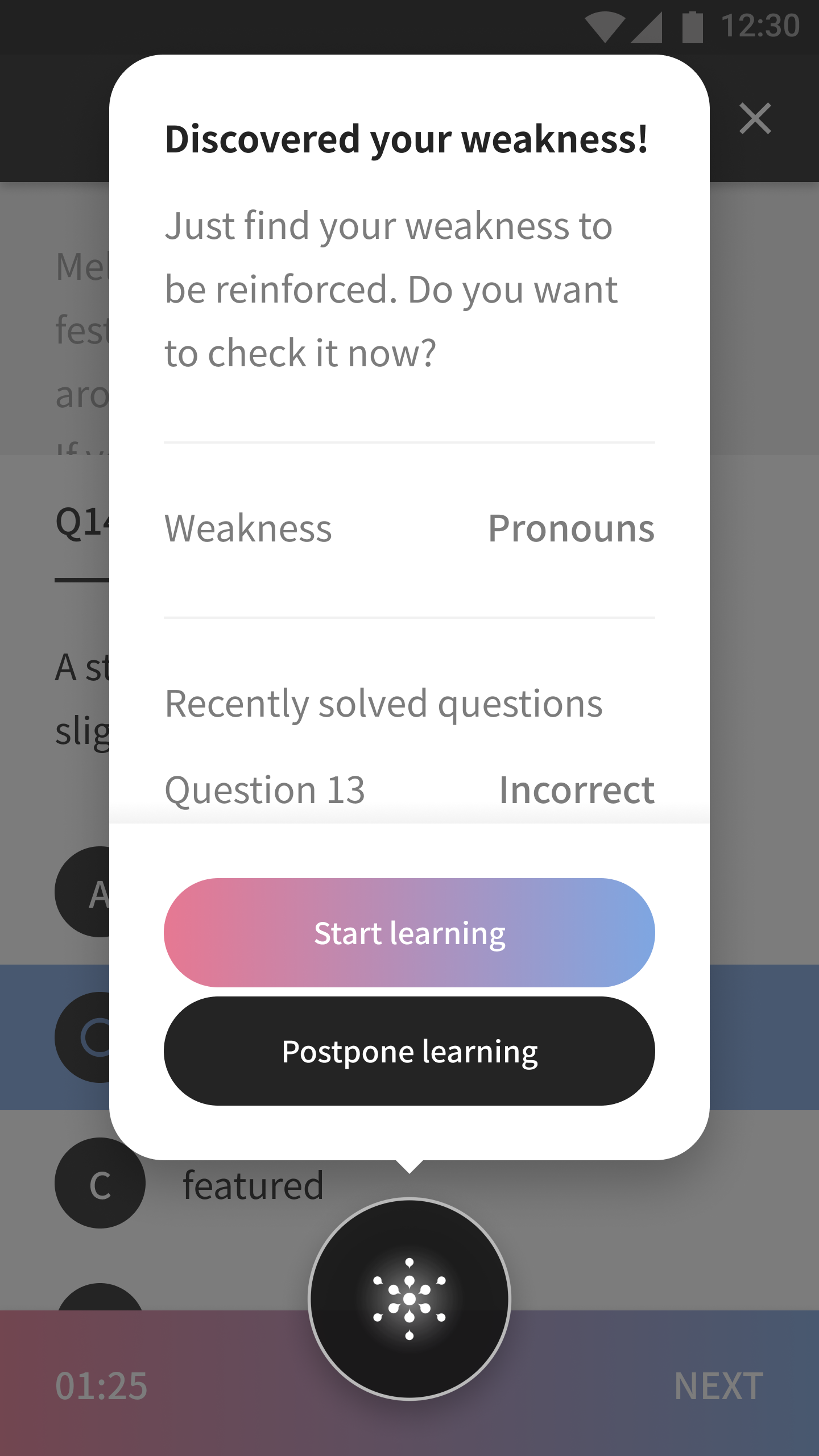}}
    \caption{Review System of \emph{Santa}}
    \label{fig:review}
\end{figure}

\subsection{Setup}
\subsubsection{Assessment Model}
Our assessment model consists of two encoder blocks ($N = 2$) with a latent space dimension of 256 ($d_{model} = 256$).
The model takes 100 interactions as input.
For pre-training, the model parameters are first randomly initialized and trained with 0.6 making rate $M$.
Then we replace the last layer of the pre-trained model with a linear layer appropriate for each label-scarce educational downstream task, and fine-tune the whole parameters of the model.
In fine-tuning, to alleviate label-scarce problem, we apply the following data augmentation strategy.
Given the original interaction sequence with a label, we select each entry in the sequence with 50\% probability to generate a subsequence with the same label.
The model is fine-tuned on these subsequences.
In both pre-training and fine-tuning, dropout rate and batch size are set to 0.2 and 128, respectively.
We use the Adam optimizer \cite{kingma2014adam} with $lr=0.001, \beta_1=0.9, \beta_2=0.999, epsilon=1e-8$, and Noam scheme \cite{vaswani_2017} to schedule the learning rate with 4000 warmup-steps.
We conduct 5-fold cross validation to select the model with the best result on the validation set, and report the evaluation result of the model on the test set.

We compare the effectiveness of the assessment model with the following content-based pre-training methods.
Since the existing content-based pre-training methods learn embedding vectors of each exercise, we replace the embedding of $\emph{exercise\_id}$ in our model with respective exercise embedding for fine-tuning.
\begin{itemize}
\item \emph{Word2Vec} \cite{mikolov2013distributedWORD2VEC} is a standard word embedding model used in many tasks.
\cite{liu_2019} used word embedding vectors obtained from Word2Vec model to generate exercise embedding vectors.
In our experiment, we obtain 256 dimension word embedding vectors from the continuous bag-of-words Word2Vec model trained on text corpus consists of exercise descriptions.
The embedding vector assigned to each exercise is the average of embedding vectors of all words appearing in the exercise description.
\item \emph{BERT} \cite{devlin2018bert} is a Transformer encoder-based bidirectional representation learning model used in various domains.
The original BERT was pre-trained by masked language modeling and next sentence prediction objectives.
However, several subsequent works \cite{liu2019roberta,lan2019albert} questioned the necessity of the next sentence prediction objective.
Accordingly, we train 2 layers and 256 dimension BERT model using only masked language modeling objective on exercise descriptions.
Similar to the case in Word2Vec, we obtain each exercise embedding vector by averaging the embedding vectors of words in the exercise description.
\item \emph{QuesNet} \cite{yin2019quesnet} is a content-based pre-training method learning unified representation for each exercise comprises of text, images and side information.
Following the method suggested in the original paper, we train bi-directional LSTM followed by multi-headed self-attention model with holed language modeling and domain oriented objectives.
The embedding vector of each exercise is computed as the sentence layer output of the model when the exercise is taken as input.
\end{itemize}

\subsubsection{Metrics}
We use the following evaluation metrics to evaluate model performance on each label-scarce educational downstream task.
For exam score prediction, we compute the Mean Absolute Error (MAE), the average of differences between the predicted exam scores and the true values. 
For review correctness prediction, we use Area Under the receiving operating characteristic Curve (AUC).

\subsection{Experimental Results}

\subsubsection{Effect of Pre-training Tasks}
We demonstrate the importance of choosing appropriate pre-training tasks by comparing models pre-trained to predict one of \emph{correctness}, \emph{correctness} + \emph{elapsed\_time}, \emph{correctness} + \emph{timeliness}, and \emph{correctness} + \emph{inactive\_time}.
Since \emph{correctness} is the feature with the most pedagogical aspect, we include \emph{correctness} in the prediction targets of all the pre-training tasks.
The results are shown in Table \ref{tab:pretask}.
For both exam score and review correctness prediction, the results show that it is the best to pre-train the model to predict \emph{correctness} and \emph{timeliness}.
This is because the exam score and review correctness distribution are related to not only correctness but also timeliness of the student's answers.
This shows the importance of choosing appropriate pre-training tasks relevant to the downstream tasks.

\begin{table}[h]
\centering
\caption{Comparison of Assessment Modeling with different pre-training tasks.}
\begin{tabular}{ccc}
\toprule
                                            & Exam score prediction & Review correctness prediction \\
\midrule
\emph{correctness}                          & $59.48\pm1.56$            & $0.852\pm0.006$ \\
\emph{correctness} + \emph{elapsed\_time}   & $59.79\pm1.49$            & $0.854\pm0.009$ \\
\emph{correctness} + \emph{timeliness}      & $\textbf{57.83}\pm1.36$   & $\textbf{0.857}\pm0.011$ \\
\emph{correctness} + \emph{inactive\_time}  & $59.73\pm1.58$            & $0.850\pm0.009$ \\
\bottomrule
\end{tabular}
\label{tab:pretask}
\end{table}

\subsubsection{Effect of Assessment Modeling}
Experimental results comparing Assessment Modeling with three different content-based pre-training methods and the model without pre-training are shown in the Table \ref{tab:expres}.
In all downstream tasks, Assessment Modeling outperforms the other methods.
Compared to the model without pre-training, MAE for exam score prediction is reduced by 21.96\% and AUC for review correctness prediction is increased by 6.99\%.
Also, Assessment Modeling improves 13.34\% MAE for exam score and 4.26\% AUC for review correctness prediction, respectively, from the previous state-of-the-art content-based pre-training methods.
These results support our claim that Assessment Modeling is more suitable for label-scarce educational downstream tasks than content-based pre-training methods.

\begin{table}[h]
\centering
\caption{Comparison to content-based pre-trainig methods.}
\begin{tabular}{cccc}
\toprule
                    & Exam score prediction & Review correctness prediction \\ 
\midrule
Without pre-train   & $74.10\pm3.76$            & $0.801\pm0.019$ \\
Word2Vec            & $68.87\pm3.20$            & $0.822\pm0.014$ \\
BERT                & $66.73\pm2.10$            & $0.814\pm0.008$ \\ 
QuesNet             & $75.44\pm1.45$            & $0.817\pm0.012$ \\
\hline
Assessment Modeling & $\textbf{57.83}\pm1.36$   & $\textbf{0.857}\pm0.011$ \\ 
\bottomrule
\end{tabular}
\label{tab:expres}
\end{table}

\section{Discussions}

\subsection{Choosing Appropriate Pre-training Task}
Determining which pre-training task is appropriate for a specific downstream task is not clear and somewhat ambiguous.
For instance, while the experimental results of including \emph{inactive\_time} in the prediction targets of pre-training shown in Table \ref{tab:pretask} is aligned with our intuition that \emph{inactive\_time} can not be an assessment, the experimental results of using \emph{timeliness} and \emph{elapsed\_time} as assessments are not.
As described in Example \ref{ex:am}, intuitively, both \emph{timeliness} and \emph{elapsed\_time} are assessments since the amount of time a student takes to respond to each exercise serves as a pedagogical evaluation.
Also, since \emph{timeliness} is a fine-grained version that binarizes \emph{elapsed time} using the time limit for each exercise, \emph{timeliness} and \emph{elapsed time} have a lot of overlap of information.
Nevertheless, according to the results shown in Table \ref{tab:pretask}, when \emph{correctness} + \emph{timeliness} (resp. \emph{correctness} + \emph{elapsed\_time}) are used as the prediction target of pre-training, there was a 2.77\% reduction (resp. 0.53\% increase) in MAE in exam score prediction than when only \emph{correctness} is used as the prediction target, so the opposite results are obtained.
These results represent the difficulty in defining appropriate pre-trainig task for Assessment Modeling.
Although there remains room for development and challenges for identifying pre-train/fine-tune relations in Assessment Modeling, we do not dig into it any further and leave it as a future work.

\subsection{Asymmetry of Assessment Modeling}
\begin{figure*}
    \centering
    \includegraphics[width=1\textwidth]{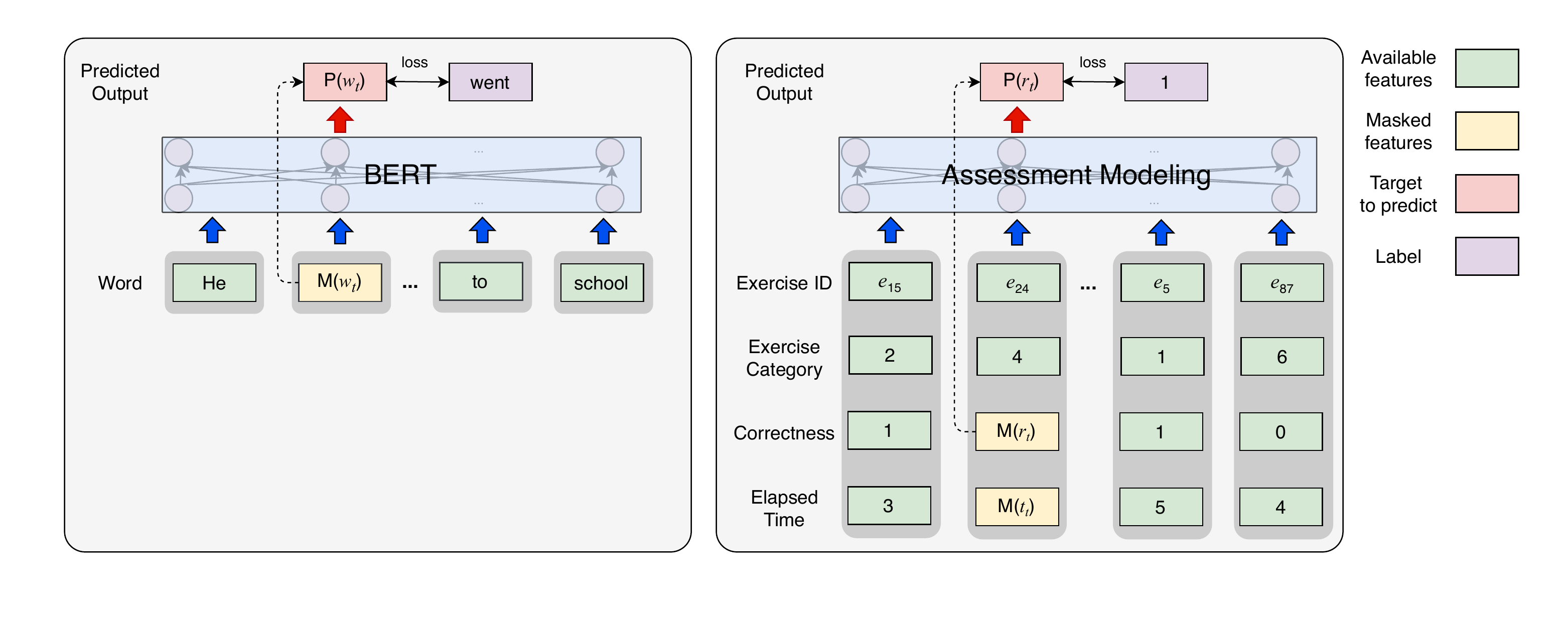}
    \caption{The issue of asymmetry arising in Assessment Modeling.
    In BERT's masked language modeling training scheme, features that are available, being masked and being predicted for each time step are all of the same nature.
    However, in Assessment Modeling, features being predicted can be a subset of features being masked, and features being masked can be a subset of features available in a specific time step.}
    \label{fig:bert_am}
\end{figure*}

While the masking scheme for Assessment Modeling was inspired by masked language modeling proposed in \cite{devlin2018bert}, there is a key difference between the two approaches (Figure \ref{fig:bert_am}).
In masked language modeling, the features available at a timestep are (the embeddings of) each word, the masked feature is the word at the timestep, and the target to predict is also the word at the timestep.
That is, there is a symmetry in that the features that are available, the features being masked, and the features being predicted are all of the same nature. 
But that is not necessarily the case in Assessment Modeling.
For example, suppose the features available at a given timestep are \emph{exercise\_id}, \emph{exercise\_category}, \emph{correctness}, and \emph{elapsed\_time}.
In the above situation, Assessment Modeling pre-training scheme may mask \emph{correctness} and \emph{elapsed\_time}, and predict just \emph{correctness}.
This asymmetry raises the issue of precisely which features to mask and which features to predict, and the choices made will have to reflect the specific label-scarce educational downstream task that Assessment Modeling is being used to prepare for.
While we draw attention to this issue, it is outside the scope of this paper and we leave the details for future study.

\section{Conclusion}
In this paper, we introduced Assessment Modeling, a class of fundamental pre-training tasks for IESs.
Our experiments show the effectiveness of Assessment Modeling as pre-training tasks for label-scarce educational problems including exam score and review correctness prediction.
Avenues of future research include 1) investigating forests of pre-train/fine-tune relations in AIEd, and 2) pre-training a model to learn not only assessments, but also representations of the contents of learning items.

\bibliographystyle{ACM-Reference-Format}
\bibliography{ref}

\end{document}